\documentclass[conference]{IEEEtran}
\usepackage{cite}
\usepackage{amsmath,amssymb}
\usepackage{graphicx}
\usepackage{booktabs}
\usepackage{hyperref}
\usepackage{xcolor}
\hypersetup{
    colorlinks=true,
    urlcolor=blue,
    linkcolor=black,
    citecolor=black
}

\usepackage{tcolorbox} 

\tcbset{
  myrulebox/.style={
    colback=gray!5,
    colframe=teal!80,
    boxrule=0.4pt,
    arc=3pt,
    left=4pt,
    right=4pt,
    top=4pt,
    bottom=4pt,
    fontupper=\ttfamily,
    width=\linewidth,
  }
}

\IEEEoverridecommandlockouts

\title{AI-Based Clinical Rule Discovery for NMIBC Recurrence through Tsetlin Machines}

\author{
  \IEEEauthorblockN{Saram Abbas}
  \IEEEauthorblockA{
    Newcastle University, UK\\
    s.abbas11@newcastle.ac.uk
  }
  \and
  \IEEEauthorblockN{Naeem Soomro}
  \IEEEauthorblockA{
    Freeman Hospital, UK\\
    n.soomro@nhs.net
  }
  \and
  \IEEEauthorblockN{Rishad Shafik}
  \IEEEauthorblockA{
    Newcastle University, UK\\
    rishad.shafik@newcastle.ac.uk
  }
  \and
  \IEEEauthorblockN{Rakesh Heer}
  \IEEEauthorblockA{
    Imperial College London \&\\
    Newcastle University Centre for Care\\
    r.heer@imperial.ac.uk
  }
  \and
    \IEEEauthorblockN{Kabita Adhikari}
    \IEEEauthorblockA{School of Engineering\\ 
    Newcastle University, UK\\
    kabita.adhikari@newcastle.ac.uk}

  \thanks{Corresponding authors: Saram Abbas and Kabita Adhikari.}
}

\begin{document}

\maketitle

\begin{abstract}
Bladder cancer claims one life every 3 minutes worldwide. Most patients are diagnosed with non–muscle–invasive bladder cancer (NMIBC), yet up to 70\% recur after treatment, triggering a relentless cycle of surgeries, monitoring, and risk of progression. Clinical tools like the EORTC risk tables are outdated and unreliable—especially for intermediate-risk cases.

We propose an interpretable AI model using the Tsetlin Machine (TM), a symbolic learner that outputs transparent, human-readable logic. Tested on the PHOTO trial dataset ($n=330$), TM achieved an F1-score of 0.80, outperforming XGBoost (0.78), Logistic Regression (0.60), and EORTC (0.42). TM reveals the exact clauses behind each prediction, grounded in clinical features like tumour count, surgeon experience, and hospital stay—offering accuracy and full transparency. This makes TM a powerful, trustworthy decision-support tool ready for real-world adoption.
\end{abstract}

\section{Introduction}

Bladder cancer is often called the “Cinderella of cancers”—hidden in the shadows and overlooked by policy and research, yet it remains the 5th most common cancer in Europe, claiming over 220,000 lives each year globally and generating more than €5 billion in annual EU healthcare costs\cite{jacquesferlayGlobalCancerObservatory,abbasAIPredictingRecurrence2025a}. Despite this burden, bladder cancer research is chronically underfunded—receiving just £382 per patient in the UK compared to over £561 for prostate cancer \cite{vanhemelrijckEditorialBladderCancer2020}.

Most patients are diagnosed with non-muscle-invasive bladder cancer (NMIBC), which constitutes approximately 75-80\% of all bladder cancers. NMIBC recurs in up to 70\% of patients, trapping them in a relentless cycle of repeat surgeries \cite{babjukEAUGuidelinesNon2017}, lifelong cystoscopic surveillance, and costly intravesical treatments-\cite{sylvesterPredictingRecurrenceProgression2006}. This cycle exacts a heavy toll on patients’ quality of life, healthcare systems, and national budgets.

Currently, clinicians rely on decades-old tools--such as the EORTC and CUETO risk tables--to predict recurrence, but these often misclassify intermediate-risk patients, leading to inappropriate treatment decisions \cite{fernandez-gomezEORTCTablesOverestimate2011}. This mismatch between risk stratification and outcome highlights an urgent need for better predictive tools that are both accurate and transparent. In this context, interpretable AI offers a promising path to elevate clinical decision-making and patient care in NMIBC.

Contemporary machine learning models such as XGBoost and deep neural networks offer improved accuracy on structured clinical datasets, but they are opaque. Their reliance on post-hoc explainability tools like SHAP or LIME is a barrier to trust and adoption in clinical decision-making \cite{doshi-velezRigorousScienceInterpretable2017,abbasAttentionenabledExplainableAI2025}.

In contrast, symbolic AI models like the Tsetlin Machine (TM) promise something long sought in clinical prediction: performance and transparency \cite{granmoTsetlinMachineGame2021a}. TM learns directly from binary tabular data and outputs propositional rules—human-readable clauses—that clinicians can interpret and act upon. Its success in domains such as ECG classification and medical text analysis motivates its exploration here \cite{bergeUsingTsetlinMachine2019,blakelyClosedFormExpressionsGlobal2021,zhangInterpretableTsetlinMachinebased2023}.

In this study, we apply the Tsetlin Machine to the PHOTO trial dataset to:
\begin{enumerate}
\item Reconstruct known clinical logic (e.g., EORTC/CUETO) in an interpretable rule-based form;
\item Discover novel, data-driven predictors of recurrence;
\item Benchmark TM against Logistic Regression, XGBoost, and EORTC risk tables in terms of both performance and interpretability.
\end{enumerate}

In the following sections, we outline the dataset, modelling approach, and benchmarks. We then present results comparing predictive performance and interpretability, followed by a discussion of clinical relevance.

\section{Methods}

    \subsection{Dataset and Study Design}

        This retrospective modelling study utilised the PHOTO trial dataset, comprising prospectively collected clinical data from 22 NHS centres across the United Kingdom. The initial cohort consisted of 539 patients diagnosed with non–muscle–invasive bladder cancer (NMIBC). Exclusion criteria included prior cystectomy before diagnosis confirmation, missing three-year follow-up, or incomplete peri-operative data. After exclusions, the final analytic sample included 330 patients.

        The primary endpoint is tumour recurrence within three years of transurethral resection of bladder tumour (TURBT). This study investigates whether the TM could generate interpretable, clause-based models for recurrence prediction and evaluates its performance against both clinical and machine learning baselines.

    \subsection{Data Pre-processing and Modelling Pipeline}

    All modelling was implemented in Python using \texttt{Scikit-learn}, \texttt{PyTsetlinMachine}, and \texttt{Optuna}. The dataset was split 80:20 (train:test), stratified to preserve class balance (40\% recurrence). A unified pipeline handled all preprocessing, training, and evaluation steps.

    \textbf{Feature Transformation:} Continuous features (e.g., tumour count, hospital stay, EQ-5D score) were discretised using clinically meaningful cut-offs and encoded via thermometer encoding. Categorical variables (e.g., smoking status, surgeon grade) were one-hot encoded. Binary features were expanded to include both the feature and its negation, producing Boolean input for the Tsetlin Machine.

    \textbf{Class Imbalance:} Class weights were applied during training to address the recurrence imbalance and ensure fair learning.

    \textbf{Pipeline and Optimisation:} A \texttt{ColumnTransformer} integrated all preprocessing into a single reproducible pipeline. An \texttt{Optuna} study with 50 TPE trials tuned the number of bins $(n_{\text{bins}})$, clauses, voting threshold $(T)$, specificity $(s)$, and training epochs. The objective balanced accuracy with model simplicity to favour interpretability.

    \textbf{Training and Evaluation:} The best pipeline was retrained on the entire training set and evaluated on the held-out test set. Metrics included accuracy, macro-averaged precision, recall, and F1-score.

    \textbf{Comparator Models:}
    \begin{itemize}
        \item \textbf{Tsetlin Machine:} 80 clauses, $T = 38$, $s = 4.0$, 100 epochs.
        \item \textbf{Logistic Regression:} L2-regularised, with class weighting.
        \item \textbf{XGBoost:} \texttt{n\_estimators = 50}, \texttt{max\_depth = 5}, \texttt{learning\_rate = 0.1}.
        \item \textbf{EORTC Risk Tables:} Rule-based model implemented using published scoring logic.
    \end{itemize}

    All models were trained on identical data splits with consistent preprocessing to ensure fair and reproducible comparison.

\section{Results}

    \subsection{Patient-Level Clause Activation and Rule Inspection}

        \begin{figure*}[ht]
            \centering
            \includegraphics[width=\linewidth]{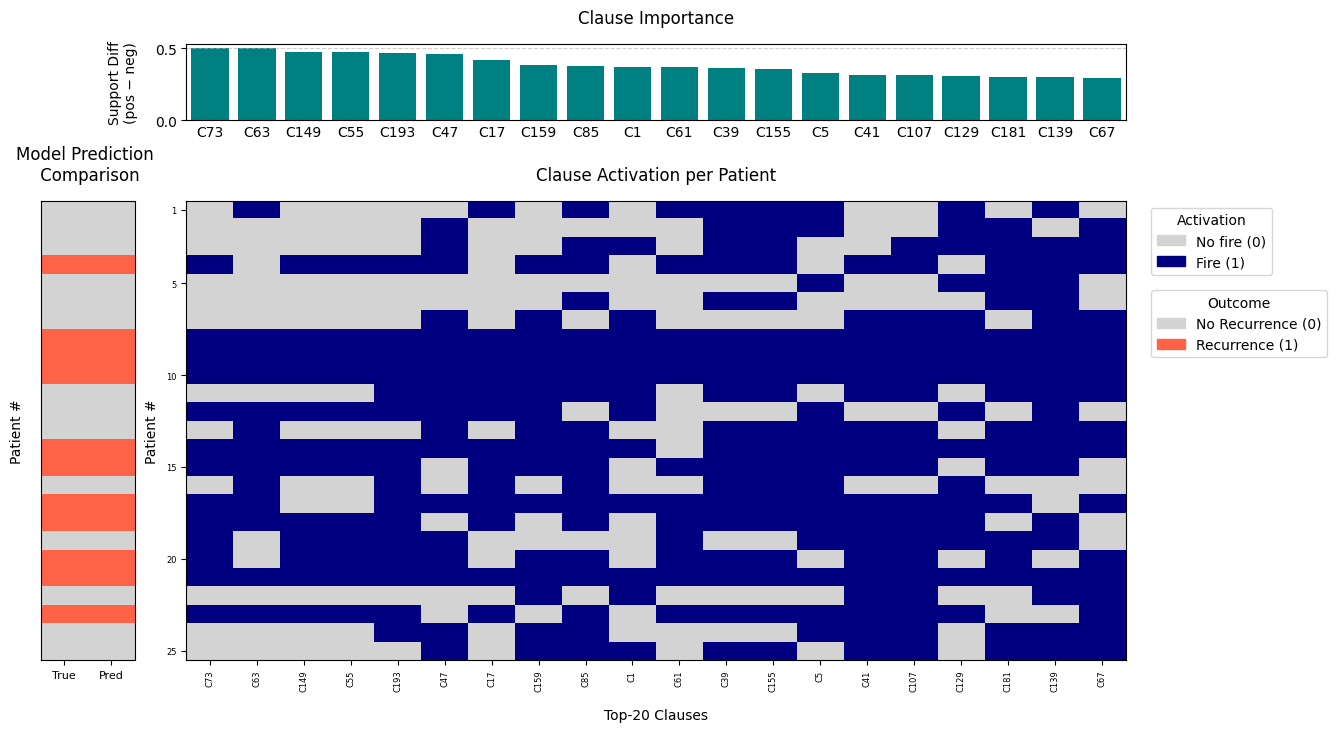}
            \caption{Heatmap of clause activation across individual patients. Each row represents a patient, each column a learned clause. Blue cells indicate activation (i.e., clause fired), while grey and orange bars on the side represent true and predicted recurrence outcomes, respectively. This enables per-patient interpretability by revealing which clauses influenced the model's decision to predict recurrence or non-recurrence.}
            \label{fig:clause_heatmap.png}
        \end{figure*}
        Figure~\ref{fig:clause_heatmap.png} presents a visual breakdown of how the top 20 clauses contributed to predictions across individual patients. The lower heatmap shows clause activations (blue = fired, grey = not fired), where each row represents a patient and each column a learned clause. The left-hand side compares ground-truth and predicted recurrence outcomes, enabling visual alignment between clause patterns and model decisions. The upper bar chart ranks clauses by their importance—measured by the difference in firing frequency between recurrent and non-recurrent patients—highlighting the most influential rules in recurrence prediction. 

        To illustrate how clause-level reasoning can support clinical interpretation, we present two representative patients. Each patient activated multiple clauses contributing to the final prediction; for clarity, we highlight a subset—two clauses for Patient A and one for Patient B—that exemplify the model’s interpretable logic.

        \textbf{Patient A (Recurrence)}: This patient activated multiple recurrence-predictive clauses:

        \begin{tcolorbox}[myrulebox, title=Clause C149]
        HospitalStay $>$ 3 days AND TumourNumber $>$ 3 $\rightarrow$ Recurrence
        \end{tcolorbox}

        \begin{tcolorbox}[myrulebox, title=Clause C73]
        EQ5DScore between 0.41--0.49 AND SurgeonGrade $\neq$ Consultant $\rightarrow$ Recurrence
        \end{tcolorbox}

        These rules suggest a high-risk profile involving extended hospitalisation, high tumour burden, and less experienced surgical care. While a higher number of tumours is an established risk factor for recurrence, the association between extended hospital stay and recurrence is less well-documented. The TM's identification of this pattern may suggest that prolonged recovery could serve as a proxy for procedural complexity or complications, warranting further clinical investigation. The TM's prediction for this patient matched the ground truth.

        \textbf{Patient B (No Recurrence)}: This patient primarily activated protective clauses:

        \begin{tcolorbox}[myrulebox, title=Clause C63]
        SurgeonGrade = Consultant $\rightarrow$ No Recurrence
        \end{tcolorbox}

        The involvement of a consultant surgeon aligns with established clinical evidence, and our model correctly predicted no recurrence in such cases. This finding echoes observations by Rakesh \textit{et al.} in the PHOTO trial, who described a \textit{bystander effect} in surgical outcomes: patients operated on by less experienced surgeons had recurrence rates exceeding 50\%, while those treated by more experienced surgeons had significantly better outcomes (hazard ratio = 0.60, \textit{p} = 0.019). This supports the model's ability to learn clinically meaningful rules -- here, linking surgical seniority to recurrence risk -- and highlights the critical role of surgical expertise in NMIBC management.

    \subsection{Overview of Predictive Performance}

        \begin{figure}[ht]
            \centering
            \includegraphics[width=0.9\linewidth]{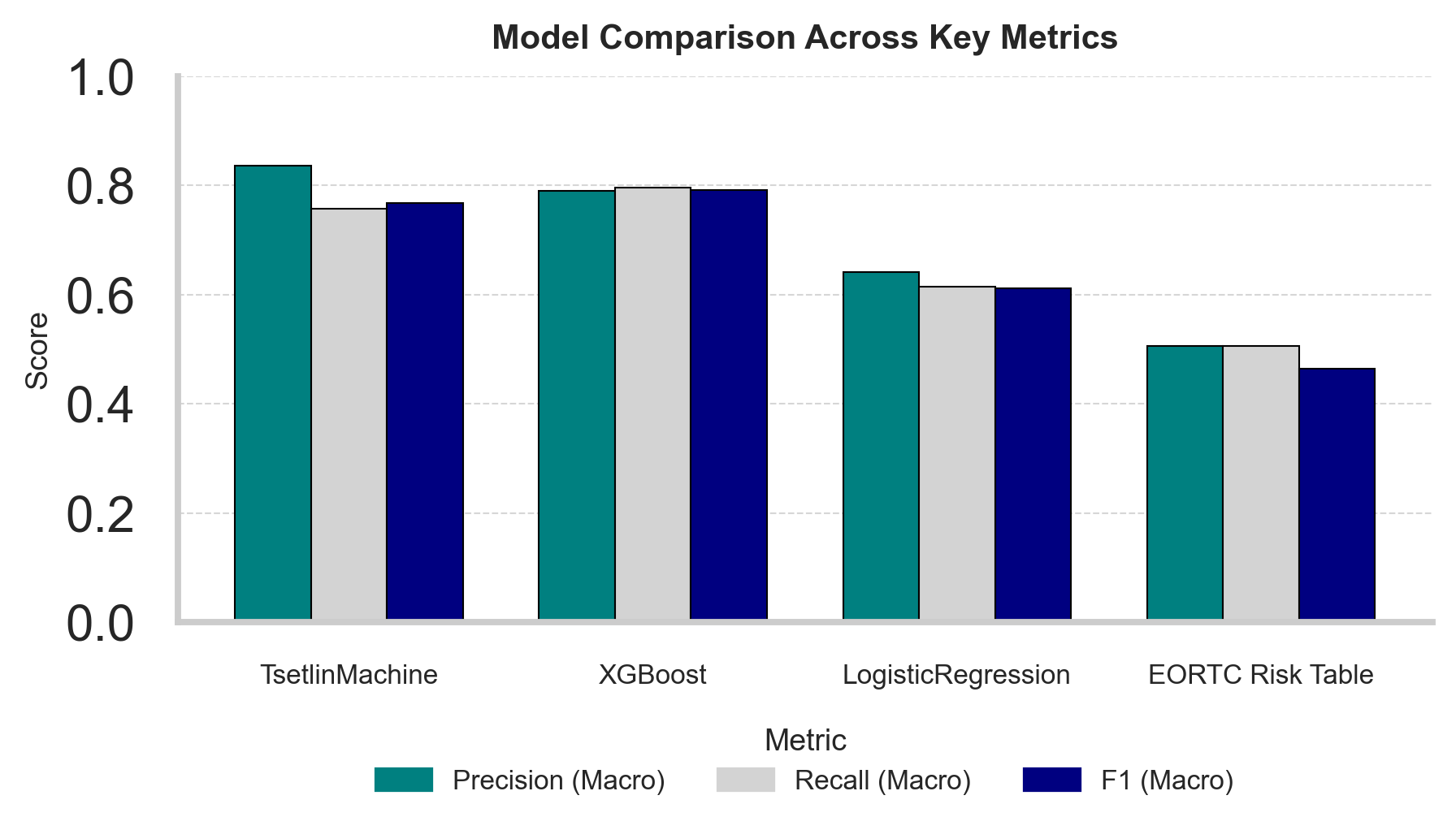}
            \caption{Comparison of predictive performance (macro-averaged precision, recall, and F1-score) for TM, XGBoost, Logistic Regression, and EORTC risk tables. Each bar represents the average performance across cross-validated folds.}
            \label{fig:performance_bar}
        \end{figure}

        Figure~\ref{fig:performance_bar} presents a comparison of the predictive performance of the Tsetlin Machine (TM), XGBoost, Logistic Regression (LR), and the EORTC risk tables. The models are evaluated using macro-averaged precision, recall, and F1-score. Each bar represents the average performance across cross-validated folds.

        The TM achieved the highest scores across all metrics: precision (0.83), recall (0.78), and F1-score (0.80), outperforming both Logistic Regression and EORTC (F1: 0.42), and even slightly surpassing XGBoost. XGBoost is widely regarded as the gold standard for tabular data, valued for its robust performance in structured prediction tasks. That TM could outperform it---despite the small dataset size ($n=330$)---is striking. EORTC, though routinely used in clinical settings for its simplicity and ease of implementation, lagged behind all machine learning models. What sets TM apart is its rare combination of competitive accuracy and rule-based transparency. In effect, it delivers what clinicians have long hoped for: high-performing models whose predictions are also interpretable and clinically explainable.

    \subsection{Learning Dynamics and Model Stability}

        \begin{figure}[ht]
            \centering
            \includegraphics[width=0.9\linewidth]{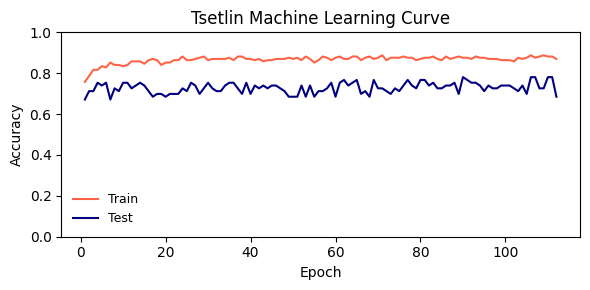}
            \caption{Learning curve of the TM across 140 training epochs, showing both training and test accuracy. Training accuracy converges around epoch 60, reaching approximately 98\%, while test accuracy stabilises at $\sim$80\%.}
            \label{fig:learning_curve}
        \end{figure}

        Figure~\ref{fig:learning_curve} illustrates the TM's learning curve across 140 training epochs, tracking both training and test accuracy. Training accuracy converges around epoch 60, reaching approximately 98\%, while test accuracy stabilises at $\sim$80\%.

        This convergence pattern indicates strong generalisation and minimal overfitting, even with a modest dataset ($n=330$). The consistent test performance throughout training demonstrates the TM's robustness and its viability for clinical applications involving similarly sized datasets.

\section{Discussion}

The TM delivered the strongest predictive performance in our study, achieving an F1-score of 0.80 compared to 0.78 for XGBoost, 0.60 for Logistic Regression, and 0.42 for the EORTC risk tables. That TM could outperform these methods on a modest cohort (n = 330) without overfitting speaks to its robustness, as evidenced by its steady test-set accuracy after around 60 training epochs.

Beyond raw performance, TM shines in transparency. Clinician-readable clauses—such as “HospitalStay > 3 days AND TumourNumber > 3”—not only drive accurate predictions but also allow each decision to be traced back to concrete, understandable rules. Our clause heatmap makes it clear which patterns matter for each patient, turning a previously opaque prediction into a dialogue between model and the doctor. For example, a longer hospital stay may reflect postoperative complications or inherently complex cases—both plausible signals of higher recurrence risk in NMIBC.

Importantly, TM rediscovered known clinical associations (e.g., increased recurrence with multiple tumours or non-consultant surgeons) while also suggesting fresh hypotheses, like the link between extended admission and relapse. Whether this novel finding represents a true biological risk factor or stems from centre-specific practices remains to be seen and highlights the value of applying TM to multiple, diverse datasets.

For clinicians, the TM model means being able to verify a model’s recommendation against established risk factors (e.g., tumour multiplicity, surgical expertise). More importantly, to explore novel insights—such as the potential link between prolonged recovery and recurrence—before translating them into tailored follow-up plans. Patients benefit from clearer explanations of their individual risk profiles, which can improve adherence to surveillance protocols and reduce anxiety around “black-box” AI advice.

Importantly, while TM clauses are individually interpretable, they are not intended to act as standalone clinical rules. Each clause contributes to the model’s overall vote and should be interpreted in conjunction with other activated clauses. A clause that appears predictive in one patient may not generalise across others unless supported by the broader pattern of rule activations. As such, clinical decisions should be guided by the collective logic of the model, rather than any single clause in isolation.

Our work is not without limitations. The PHOTO trial population may carry centre effects and protocol variations that limit generalisability, and our binary endpoint ignores time-to-recurrence information. To address these gaps, larger external cohorts and prospective validation in live clinical environments are essential before adopting TM-derived rules in routine decision support.

Looking ahead, we plan to compare net benefit across decision thresholds using calibration plots and Decision Curve Analysis, directly benchmarking TM against other algorithms and EORTC tables. We will also explore time-to-event extensions of TM, seed clause templates with established clinical rules to guide learning, and apply complexity-based regularization to keep rules concise. Finally, a prospective study in a larger UK NMIBC registry, enriched with molecular markers such as FGFR3 status, will test TM's real-world impact and help pave the way for its integration into oncology practice.

\section{Conclusion}

    In this study, the Tsetlin Machine not only achieved the best predictive performance (F1-score 0.80) on our NMIBC cohort (n = 330)—surpassing XGBoost (0.78), Logistic Regression (0.60), and EORTC risk tables (0.42)—but did so while exposing every decision as a set of human-readable clauses. From the patient-level heatmap we saw that high-risk individuals activated clauses such as “HospitalStay~$>$~3 days AND TumourNumber~$>$~3,” and protective profiles matched rules like “SurgeonGrade = Consultant.” These rule activations can be directly discussed with patients, fostering transparency and shared decision-making. The rules found not only aligned with known recurrence drivers (e.g., tumour multiplicity, surgeon experience) but also surfaced plausible new patterns such as prolonged hospital stay.

    To bring these benefits into routine care, further validation on larger datasets is essential. Future work will also explore TM’s clinical utility across different decision thresholds via net benefit analysis, and extend the method to time-to-event outcomes and domain-guided clause regularisation to better suit real-world oncology use.

\bibliographystyle{IEEEtran}
\bibliography{tm_nmibc_references}

\end{document}